\documentclass[runningheads]{llncs}
\usepackage{graphicx}

\usepackage{tikz}
\usepackage{comment} 
\usepackage{amsmath,amssymb} 
\usepackage{color}
\usepackage{epsfig}
\usepackage{hyperref}
\usepackage{booktabs}

\begin{document}
\pagestyle{headings}
\mainmatter

\title{A Unified Framework of Surrogate Loss by Refactoring and Interpolation}

\author{Lanlan Liu\inst{1, 2} \and
Mingzhe Wang\inst{2} \and
Jia Deng\inst{2}}
\authorrunning{L. Liu et al.}
\institute{University of Michigan, Ann Arbor MI 48105, USA \\
\email{llanlan@umich.edu}\\
\and Princeton University, Princeton NJ 08544, USA \\
\email{\{mingzhew,jiadeng\}@cs.princeton.edu}}
\maketitle

\begin{abstract}
We introduce UniLoss, a unified framework to generate surrogate losses for training deep networks with gradient descent, reducing the amount of manual design of task-specific surrogate losses. Our key observation is that in many cases, evaluating a model with a performance metric on a batch of examples can be refactored into four steps: from input to real-valued scores, from scores to comparisons of pairs of scores, from comparisons to binary variables, and from binary variables to the final performance metric. Using this refactoring we generate differentiable approximations for each non-differentiable step through interpolation. Using UniLoss, we can optimize for different tasks and metrics using one unified framework, achieving comparable performance compared with task-specific losses. We validate the effectiveness of UniLoss on three tasks and four datasets. Code is available at \url{https://github.com/princeton-vl/uniloss}.
\end{abstract}

\section{Introduction}
\label{sec:intro}
Many supervised learning tasks involve designing and optimizing a loss function
that is often different from the actual performance metric for evaluating models. 
For example, cross-entropy is a popular loss function for training a multi-class
classifier, but when it comes to comparing models on a test set, 
classification error is used instead. 

Why not optimize the performance metric directly? Because 
many metrics or output decoders are non-differentiable and cannot be optimized with gradient-based methods
such as stochastic gradient descent. Non-differentiability occurs when 
the output of the task is discrete (e.g. class labels), or when the output is continuous
but the performance metric has discrete operations (e.g. percentage of real-valued predictions
within a certain range of the ground truth).

To address this issue, designing a differentiable
loss that serves as a surrogate to the original metric is standard practice. 
For standard tasks with simple output and metrics, there exist
well-studied surrogate losses. For
example, cross-entropy or hinge loss for classification, both of
which have proven to work well in practice.

However, designing surrogate losses can sometimes incur substantial manual effort, including a large amount of
trial and error and hyper-parameter tuning. 
For example, a standard evaluation of single-person human pose estimation---predicting the
2D locations of a set of body joints for a single person in an image---involves computing
the percentage of predicted body joints that are within a given radius of the ground
truth. 
This performance metric is non-differentiable. 
Existing work instead trains a deep network
to predict a heatmap for each type of body joints, minimizing the difference between the predicted
heatmap and a ``ground truth'' heatmap consisting of a Gaussian bump at the ground truth
location~\cite{tompson2014joint,newell2016stacked}.
The decision for what error function to use for comparing heatmaps and 
how to design the ``ground truth'' heatmaps
are manually tuned to optimize performance. 

This manual effort in conventional losses is tedious but necessary,
because a poorly designed loss can be misaligned with the final performance metric and
lead to ineffective training. As we show in the experiment section, without carefully-tuned loss hyper-parameters, conventional manual losses can work poorly. 

\begin{figure*}[t]
\begin{center}
\includegraphics[width=1\linewidth]{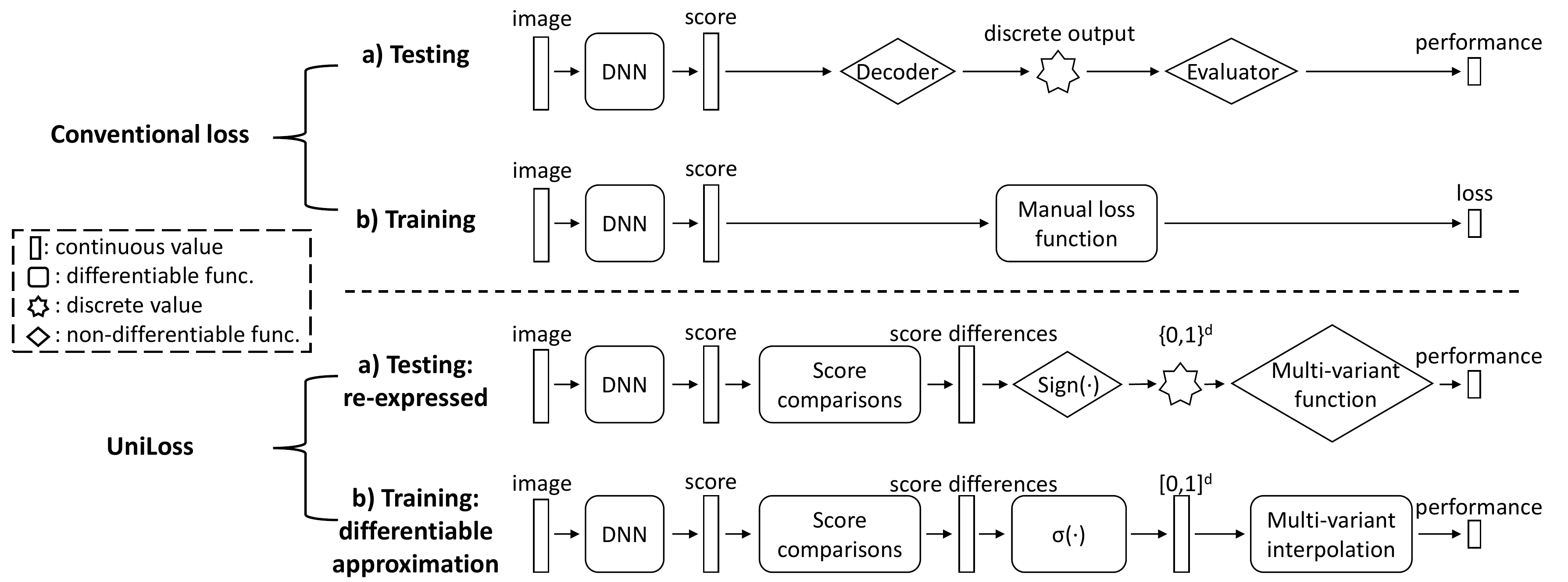}
\end{center}
 \caption{Computation graphs for conventional losses and UniLoss. Top: (a) testing for conventional losses. The decoder and evaluator are usually non-differentiable. (b) training for conventional losses. To avoid the non-differentiability, conventional methods optimize a manually-designed differentiable loss function instead during training. Bottom: (a) refactored testing in UniLoss. We refactor the testing so that the non-differentiability exists only in Sign($\cdot$) and the multi-variant function. (b) training in UniLoss with the differentiable approximation of refactored testing. $\sigma(\cdot)$ is the sigmoid function. We approximate the non-differentiable components in the refactored testing pipeline with  interpolation methods. }
\label{fig:overview}
\end{figure*}

In this paper, we seek to reduce the efforts of manual design of surrogate losses by introducing a unified surrogate loss framework applicable to a wide range of tasks.
We provide a unified framework to mechanically generate a surrogate loss given a performance metric
in the context of deep 
learning.
This means that we only need to specify the performance metric (e.g.
classification error) and the inference
algorithm---the network architecture, a ``decoder'' that converts the network output
(e.g. continuous scores) to the final
output (e.g. discrete class labels), and an ``evaluator'' that converts the labels to final metric---and the rest is taken care of as part of the 
training algorithm.

We introduce UniLoss (Fig.~\ref{fig:overview}), a unified framework to generate surrogate losses for training deep networks with gradient descent.  
We maintain the
basic algorithmic structure of mini-batch gradient descent: for each mini-batch, we perform
inference on all examples, compute a loss using the results and the ground truths, and generate
gradients using the loss to update the network parameters. Our novelty is that we generate all the surrogate losses in a unified framework for various tasks instead of manually designing it for each task.

The key insight behind UniLoss is that for many tasks and performance metrics, evaluating a
deep network on a set of training examples---pass the examples through the network, the output
decoder, and the evaluator to the performance metric---can be refactored into a sequence of four 
transformations:
the training examples are first transformed to a set of real scores, then to some real numbers
 representing comparisons (through subtractions) of certain pairs of the real-valued scores,
  then to a set of binary values representing the signs of the comparisons, and finally to a single
real number. 
Note that the four transforms do not necessarily correspond to
running the network inference, the decoder, and the evaluator. 

Take multi-class classification as an example, the training examples are first transformed to
a set of scores (one per class per example), and then to pairwise comparisons (subtractions) between the scores for each example (i.e. the argmax operation), and then to a set of binary values, and finally to a classification accuracy. 

The final performance metric is non-differentiable with respect to network weights because the decoder and the evaluator
are non-differentiable. But 
this refactoring allows us to
generate a differentiable approximation of each non-differentiable transformation through
interpolation. 

Specifically, the transformation from comparisons to binary variables is nondifferentiable, we can approximate it by using the sigmoid
function to interpolate the sign function.
And the transformation from binary variables to final metric may be nondifferentiable, we can approximate it by multivariate interpolation.

The proposed UniLoss framework is general and can be applied to various tasks and performance metrics.
Given any performance metric involving discrete operations, 
to the best of our knowledge, the discrete operations can always be refactored to step functions that first make some differentiable real-number comparisons non-differentiable, and any following operations, which fit in our framework. Example tasks include classification scenarios such as accuracy in image classification, precision and recall in object detection; 
ranking scenarios such as average precision in binary classification, area under curve in image retrieval;
pixel-wise prediction scenarios such as mean IOU in segmentation, PCKh in pose estimation.

To validate its effectiveness, we perform experiments on three representative tasks from three different scenarios. 
We show that UniLoss performs well on a classic classification setting, multi-class classification, compared with the well-established conventional losses.
We also demonstrate UniLoss's ability in a ranking scenario that evolves ranking multiple images in an evaluation set: average precision (area under the
precision-recall curve) in unbalanced binary classification. 
In addition, we experiment with pose estimation where the output is structured as pixel-wise predictions.

Our main contributions in this work are: 
\begin{itemize}
 \item We present a new  perspective of finding surrogate losses: evaluation can be refactored as a sequence of four transformations, where each nondifferentiable transformation can be tackled individually.
 \item We propose a new method: a unified framework to generate losses for various tasks reducing task-specific manual design.
 \item We validate the new perspective and the new method on three tasks and four datasets, achieving comparable performance with conventional losses.
\end{itemize}

\section{Related Work}

\subsection{Direct Loss Minimization} 

The line of direct loss minimization works is related to UniLoss because we share a similar idea of finding a good approximation of the performance metric. 
There have been many efforts to directly minimize specific classes of tasks and metrics. 

For example, \cite{taylor2008softrank} optimized ranking metrics such as Normalized Discounted
 Cumulative Gain by smoothing them with an assumed probabilistic distribution of
 documents. \cite{henderson2016end} directly optimized mean average precision in object
 detection by computing ``pseudo partial derivatives'' for various continuous variables. 
 \cite{nguyen2013algorithms} explored to optimize the 0-1 loss in binary classification by 
  search-based methods including branch and bound search, combinatorial search, and also coordinate descent on the relaxations of 0-1 losses.
  \cite{Liu2016ICML} proposed to improve the conventional cross-entropy loss by multiplying a preset constant with the angle in the inner product of the $\mathtt{softmax}$ function to encourage large margins between classes. 
  \cite{fu2018end} proposed an end-to-end optimization approach for speech enhancement by directly optimizing 
  short-time objective intelligibility (STOI)
  which is a differentiable performance metric. 
   
 In addition to the large algorithmic differences, these works also differ from ours in
 that they are tightly coupled with specific  tasks and applications. 

\cite{hazan2010direct} and \cite{song2016training} proved that under mild conditions, optimizing a max-margin
structured-output loss is asymptotically equivalent to directly optimizing the performance
metrics. 
Specifically, assume a model
in the form of a differentiable scoring function $S(x, y;w):\mathcal{X} \times \mathcal{Y}\rightarrow \mathbf{R}$ that
evaluates the compatibility of output $y$ with input $x$.
During inference, they predict the
$y_w$ with the best score:
\begin{equation}
y_w = \underset{y}{\operatorname{argmax}}~S(x,y;w).
\end{equation}
During training, in addition to this regular inference, they also
perform the \emph{loss-augmented inference}~\cite{tsochantaridis2005large,hazan2010direct}:
\begin{equation}
y^\dagger = \underset{y}{\operatorname{argmax}}~S(x,y;w) + \epsilon \xi(y_w, y), 
\label{eqn:max-direct}
\end{equation}
where $\xi$ is the final performance metric (in terms of error), and $\epsilon$ is a small time-varying
weight.
\cite{song2016training} generalized this result from linear scoring functions
to arbitrary scoring functions, and developed an efficient loss-augmented inference algorithm
to directly optimize average precision in ranking tasks. 

While above max-margin losses can ideally work with many different performance
metrics $\xi$, its main limitation in practical use is that it can be highly nontrivial to design an efficient
algorithm for the loss-augmented inference, as it often requires some
clever factorization of the performance metric $\xi$ over the components of the structured
output $y$. In fact, for many metrics the loss-augmented inference is NP-hard and one must
resort to designing approximate algorithms, which further increases the difficulty of practical use. 
 
In contrast, our method does not demand the same level of human ingenuity. The main human
effort involves refactoring the inference code and evaluation code to a particular
format, which may be further eliminated by automatic code analysis. There is no need to design a new inference algorithm over discrete outputs and analyze its efficiency. 
The difficulty of designing loss-augmented inference algorithms for each individual task makes it impractical to compare fairly with max-margin methods on diverse tasks, because it is unclear how to design the inference algorithms.

Recently, some prior works propose to directly optimize the performance metric  
by learning a parametric surrogate loss~\cite{huang2019addressing,wu2018learning,santos2017learning,josif2019learning,engilberge2019sodeep}. 
During training, the model is updated to minimize the current surrogate loss while the parametric surrogate loss is also updated to align with the performance metric. 

Compared to these methods, 
UniLoss does not involve any learnable parameters in the loss.
As a result, UniLoss can be applied universally across different settings without any training,
and the parametric surrogate loss has to be trained separately 
for different tasks and datasets.

Reinforcement Learning inspired algorithms have been used to optimize performance metrics
for structured output problems, especially those that can be formulated as taking a
sequence of actions~\cite{ranzato2015sequence,liu2017improved,caicedo2015active,yeung2016end,zhou2018improving}. For example,
\cite{liu2017improved} use policy
gradients~\cite{sutton2000policy} to optimize metrics for image captioning. 

We differ from these
approaches in two key aspects. First, we do not need to formulate a task as
a sequential decision problem, which is natural for certain tasks such as text
generation, but unnatural for others such as human pose estimation. Second, these methods treat performance metrics as black boxes, whereas we assume access to
the code of the performance metrics, which is a valid assumption in most
cases. This access allows us to reason about
the code and generate better gradients.

\subsection{Surrogate Losses}
There has been a large body of literature studying surrogate losses, for tasks including multi-class
classification~\cite{bartlett2006convexity,zhang2004statistical,tewari2007consistency,crammer2001algorithmic,allwein2000reducing,ramaswamy2013convex,ramaswamy2014consistency}, binary classification~\cite{bartlett2006convexity,zhang2004statistical,ramaswamy2013convex,ramaswamy2014consistency} and pose estimation~\cite{tompson2014joint}. 
Compared to them, UniLoss reduces the manual effort to design task-specific losses. UniLoss, as a general loss framework, can be applied to all these tasks and achieve comparable performance.

\section{UniLoss}

\subsection{Overview}
\label{sec:method_overview}
UniLoss provides a unified way to generate a surrogate loss for training deep networks with mini-batch gradient descent without task-specific design. In our general framework, we first re-formulate the evaluation process and then approximate the non-differentiable functions using interpolation.

\subsubsection{Original Formulation}
Formally, let $\mathbf{x} = (x_1, x_2, \ldots, x_n) \in \mathcal{X}^n$ be a set of $n$ training examples and
$\mathbf{y}=(y_1, y_2, \ldots, y_n) \in \mathcal{Y}^n$ be the ground truth. Let $\phi(\cdot;w):
\mathcal{X} \rightarrow \mathbf{R}^d$ be a deep network parameterized by weights $w$ that outputs a $d$-dimensional
vector; let 
$\delta:\mathbf{R}^d \rightarrow \mathcal{Y}$ be a decoder that decodes the network output to a possibly
discrete final output; let $\xi: \mathcal{Y}^n \times \mathcal{Y}^n \rightarrow \mathbf{R}$
be an evaluator. $\phi$ and $\delta$ are applied in a mini-batch fashion on $\mathbf{x} = (x_1, x_2, \ldots, x_n)$; the performance $e$ of the deep network is then
\begin{equation}
e=\xi(\delta(\phi(\mathbf{x};w)), \mathbf{y}).
\label{eqn:e_ori}
\end{equation}

\subsubsection{Refactored Formulation}
 Our approach seeks to  find a surrogate loss to minimize $e$, with
the novel observation that in many cases $e$ can be refactored as
\begin{equation}
e = g (h (f(\phi(\mathbf{x};w),\mathbf{y}) ) ),
\label{eqn:reexpr}
\end{equation}
 where $\phi(\cdot;w)$ is the same as in Eqn.~\ref{eqn:e_ori}, representing a deep neural network,
  $f: \mathbf{R}^{n\times d} \times \mathcal{Y}^n
\rightarrow \mathbf{R}^{l}$ is differentiable and maps outputted
real numbers and the ground truth to $l$ comparisons each
representing the difference between certain pair of real numbers, $h: \mathbf{R}^{l}
\rightarrow \mathbf\{0,1\}^{l} $ maps the $l$ score differences to $l$ binary variables, and 
$g:\mathbf\{0,1\}^{l} \rightarrow \mathbf{R}$ computes the performance
metric from binary variables. Note that $h$ has a restricted form that always maps continuous values to binary
values through sign function, whereas $g$ can be arbitrary
computation that maps binary values to a real number.

We give intermediate outputs some notations:
\begin{itemize}
    \item Training examples $\mathbf{x},\mathbf{y}$
are transformed to scores $\mathbf{s} = (s_1, s_2, \ldots, s_{nd})$, where
$\mathbf{s} = \phi(\mathbf{x};w)$.
    \item $\mathbf{s}$ is converted to comparisons (differences of two scores) $\mathbf{c} = (c_1, c_2, \ldots, c_l)$, where $\mathbf{c} = f(\mathbf{s}, \mathbf{y})$.
    \item $\mathbf{c}$ is converted to 
binary variables $\mathbf{b} = (b_1, b_2, \ldots, b_l)$ representing the binary outcome of the comparisons, 
where $\mathbf{b} = h(\mathbf{c})$.
    \item The binary variables are transformed to a single real number by $e=g(\mathbf{b})$.
\end{itemize}

This new refactoring of a performance metric allows us to
decompose the metric $e$ with $g$, $h$, $f$ and $\phi$, where 
$\phi$ and
$f$ are
differentiable functions but 
$h$ and $g$ are often non-differentiable.
The non-differentiability of $h$ and $g$ causes $e$ to be
non-differentiable with respect to network weights $w$.

\subsubsection{Differentiable Approximation}
Our UniLoss generates differentiable approximations of the
non-differentiable $h$ and $g$ through interpolation, thus making the metric $e$
optimizable with gradient descent.
Formally, UniLoss gives a differentiable approximation $\tilde{e}$
\begin{equation}
\tilde{e} = \tilde{g} (\tilde{h} (f( \phi(\mathbf{x};w),\mathbf{y}))),
\label{eqn:approx}
\end{equation}
where $f$ and $\phi$ are the same as in Eqn.~\ref{eqn:reexpr}, and $\tilde{h}$ and $\tilde{g}$
are the differentiable approximation of $h$ and $g$.  
We explain a concrete example of multi-class classification and introduce the refactoring and 
interpolation in detail based on this example in the following sections.

\subsection{Example: Multi-class Classification}
\label{sec:cls_example}

We take multi-class classification as an example to show how refactoring works. First, we give formal definitions of multi-class classification and the performance metric: prediction accuracy. 

Input is a mini-batch of images $\mathbf{x}=(x_1, x_2, \ldots, x_n)$ 
and their corresponding ground truth labels are $\mathbf{y}=(y_1, y_2, \ldots, y_n)$ 
where $n$ is the batch size. $y_i\in \{1,2,\ldots, p\}$ 
and $p$ is the number of classes, which happens to be the same value as $d$ in Sec.~\ref{sec:method_overview}. 
A network $\phi(\cdot;w)$ outputs a score matrix $\mathbf{s}=[s_{i,j}]_{n\times p}$ 
and $s_{i,j}$ represents the score for the i-th image belongs to the class j.

The decoder $\delta(\mathbf{s})$ decodes $\mathbf{s}$ into the discrete outputs 
$\tilde{\mathbf{y}}=(\tilde{y}_1, \tilde{y}_2, \ldots, \tilde{y}_n)$ 
by 
\begin{equation}
\label{eqn:argmax}
\tilde{y}_i= \underset{1\leq j \leq p}{\operatorname{argmax} }~s_{i,j}, 
\end{equation}
and $\tilde{y}_i$ represents the predicted label of the i-th image for $i=1,2,\ldots,n$.

The evaluator $\xi(\tilde{\mathbf{y}}, \mathbf{y})$ evaluates the accuracy $e$ from $\tilde{\mathbf{y}}$ and $\mathbf{y}$ by 
\begin{equation}
\label{eqn:compare}
e=\frac{1}{n}\sum_{i=1}^n [y_i=\tilde{y}_i],
\end{equation}
where $[\cdot]$ is the Iverson bracket. 

Considering above together,
the predicted label for an image is correct if and only if 
the score of its ground truth class is higher than the score of every other class:
\begin{equation}
\begin{split}
\label{eqn:final_pair}
[y_i=\tilde{y}_i]=\underset{j\ne y_i}{\underset{1\leq j \leq p}{\land}} [s_{i,y_i} - s_{i,j} > 0], 
\text{ for all }1 \leq i \leq n,
\end{split}
\end{equation} 
where $\land$ is  logical and.

We thus refactor the decoding and evaluation process as a sequence of  $f(\cdot)$ that transforms $\mathbf{s}$ to comparisons---$s_{i,y_i} - s_{i,j} \text{ for all } 1 \leq i \leq n, 1\leq j \leq p, \text{and } j\ne y_i$ ($n \times (p-1)$ comparisons in total), $h(\cdot)$ that transforms comparisons to binary values using $[\cdot > 0]$, and $g(\cdot)$ that transforms binary values to $e$ using logical and.
Next, we introduce how to 
refactor the above procedure into our formulation and approximate $g$ and $h$.

\subsection{Refactoring}
\label{sec:cls_decomp}

Given a performance metric, we refactor it in the form of Eqn.~\ref{eqn:reexpr}. We first transform the training images into scores
$\mathbf{s} = (s_1, s_2, \ldots, s_{nd})$.
We then get the score comparisons (differences of pairs of scores) $\mathbf{c} = (c_1, c_2, \ldots, c_l)$
using $\mathbf{c} = f(\mathbf{s},\mathbf{y})$. Each comparison is 
$c_i= s_{k_i^1} - s_{k_i^2}, 1\leq i\leq l, 1\leq k_i^1, k_i^2 \leq nd$.
The function $h$ then transforms the comparisons to 
binary values by $\mathbf{b} = h(\mathbf{c })$. $h$ is the sign function, 
i.e. 
$b_i=[c_i > 0], 1\leq i\leq l$.
The function $g$ then computes $e$ by $e=g(\mathbf{b})$,
where $g$ can be arbitrary
computation that converts binary values to a real number. In practice, $g$ can be complex
and vary significantly across tasks and metrics.

Given any performance metrics 
involving discrete operations in function $\xi$ and $\delta$ in Eqn.~\ref{eqn:e_ori}
(otherwise the metric $e$ is differentiable and trivial to be handled), 
the computation of  function $\xi(\delta(\cdot))$ 
can be refactored as a sequence of 
continuous operations (which is optional), 
 discrete operations that make some differentiable real numbers non-differentiable, 
 and any following operations.
 The discrete operations always occur when there are step functions, which can be expressed as comparing two numbers, to the best of our knowledge.

This refactoring is usually straightforward to obtain from the specification of the decoding
and evaluating procedures. The only manual effort is in identifying the discrete comparisons (binary variables). 
Then we simply write the discrete comparisons as function $f$ and $h$,
and represent its following operations as function $g$.

In later sections we will show how to identify the binary variables for three commonly-used metrics in three scenarios, which can
be easily extended to other performance metrics. 
On the other hand, this process is largely a mechanical exercise, as it is equivalent to rewriting some existing code in an alternative rigid format.

\subsection{Interpolation}
The two usually non-differentiable functions $h$ and $g$ are approximated by interpolation methods individually.

\subsubsection{Scores to Binaries: $h$.}
In $\mathbf{b} = h(\mathbf{c})$, each element $b_i = [c_i > 0]$. We approximate
the step function $[\cdot]$ using the $\mathtt{sigmoid}$ function. That is,
$\tilde{\mathbf{b}} = \tilde{h}(\mathbf{c})=(\tilde{b}_1, \tilde{b}_2, \ldots, \tilde{b}_l)$, and each element 
\begin{equation}
\tilde{b}_i = \mathtt{sigmoid}(c_i),
\label{eqn:sigmoid}
\end{equation}
where $1\leq i\leq l$. 
We now have $\tilde{h}$ as the differentiable approximation of $h$.

\subsubsection{Binaries to Performance: $g$.}
\label{sec:method_g}

We approximate $g(\cdot)$ in $e=g(\mathbf{b})$ by multivariate interpolation over the input $\mathbf{b} \in
\{0,1\}^{l}$. More specifically,
we first sample a set of 
configurations as ``anchors'' $\mathbf{a} = (\mathbf{a}_1, \mathbf{a}_2, \ldots, \mathbf{a}_t)$,
where $\mathbf{a}_i$ is a configuration of $\mathbf{b}$,
and compute the output values
$g(\mathbf{a}_1),g(\mathbf{a}_2), \ldots, g(\mathbf{a}_t)$, where
 $g(\mathbf{a}_i)$ is the actual performance metric value
and $t$ is the number of anchors sampled.

We then get an interpolated function over the anchors as 
$\tilde{g}(\cdot;\mathbf{a})$. 
We finally get
$\tilde{e} = \tilde{g}(\tilde{\mathbf{b}};\mathbf{a})$, where $\tilde{\mathbf{b}}$ is computed from $\tilde{h}$, $f$ and $\phi$.

By choosing a
differentiable interpolation method, the $\tilde{g}$ function becomes trainable using gradient-based methods.
We use a common yet effective interpolation method: inverse distance weighting (IDW)~\cite{shepard1968two}:
\begin{equation}
  \tilde{g}(\mathbf{u};\mathbf{a}) =
  \begin{cases}
  \frac{\sum_{i=1}^t{\frac{1}{d(\mathbf{u},\mathbf{a}_i)}  g(\mathbf{a}_i)}}{\sum_{i=1}^t{\frac{1}{d(\mathbf{u},\mathbf{a}_i)}}}, & d(\mathbf{u},\mathbf{a}_i)\ne 0 \text{ for }1 \leq i \leq t; \\
  g(\mathbf{a}_i), & d(\mathbf{u}, \mathbf{a}_i) = 0 \text{ for some }i. \\
  \end{cases}
\end{equation}
where $\mathbf{u}$ represents the input to $\tilde{g}$ and  $d(\mathbf{u},\mathbf{a}_i)$ is the Euclidean distance between $\mathbf{u}$ and $\mathbf{a}_i$.

We select a subset of anchors based on the
current training examples. We use a mix of three types of anchors---good anchors with high performance values globally, bad anchors with low performance
values globally, and nearby anchors that are close to the current configuration, which is computed from the current training examples and  network weights.
By using both the global information from the good and bad anchors
and the local information from the nearby anchors,
we are able to get an informative interpolation surface.
On the contrast, a naive random anchor sampling strategy does not give informative interpolation surface and cannot train the network at all in our experiments.

More specifically, we adopt a straightforward anchor sampling strategy for all tasks and metrics: we obtain good anchors by flipping some bits from the best anchor,
which is the ground truth. 
The bad anchors are generated by randomly sampling binary values. 
The nearby anchors are obtained by flipping some bits from the current configuration.

\section{Experimental Results} 
To use our general framework UniLoss on each  task, we refactor the evaluation process of the task into the format in Eqn.~\ref{eqn:reexpr}, and then approximate the non-differentiable functions $h$ and $g$ using the interpolation method in Sec. 3. 

We validate the effectiveness of the UniLoss framework in three representative tasks in different scenarios: a ranking-related task using a set-based metric---average precision, a pixel-wise prediction task, and a common classification task.
For each task, we demonstrate how to formulate the evaluation process to our refactoring and compare our UniLoss with interpolation to the conventional task-specific loss. 
More implementation details and analysis can be found in the appendix.

\subsection{Tasks and Metrics}

\subsubsection{Average Precision for Unbalanced Binary Classification}
Binary classification is to classify an example from two
classes---positives and negatives. Potential applications include face classification and image retrieval. 
It has unbalanced number of positives and negatives in most cases, which results in that a typical classification metric such as accuracy as in regular classification cannot demonstrate how good is a model properly. For example, when the positives to  negatives is 1:9, predicting all examples as negatives gets 90\% accuracy. 

On this unbalanced binary classification, other metrics such as precision, recall and average precision (AP), i.e. area under the
precision-recall curve, are more descriptive metrics. We use AP as our target metric in this task.

It is notable that AP is fundamentally different from accuracy because it is a set-based metric. It can only be evaluated on a set of images, and involves not only the correctness of each image but also the ranking of multiple images. 
This task and metric is chosen to demonstrate that UniLoss can effectively optimize for a set-based performance metric that requires ranking of the images.

\subsubsection{PCKh for Single-Person Pose Estimation}
Single-person pose estimation predicts the
localization of human joints. More specifically, given an image, it predicts the location of the joints. It is usually formulated as a pixel-wise prediction problem, where the neural network outputs a score for each pixel indicating how likely is the location can be the joint.

Following prior work, we use PCKh (Percentage of Correct Keypoints wrt to head size) as the performance metric. 
It computes the percentage of the predicted joints that are within a given radius $r$ of the ground truth. The radius is half of the head segment length.
This task and metric is chosen to validate the effectiveness of UniLoss in optimizing for a pixel-wise prediction problem.

\subsubsection{Accuracy for Multi-class Classification}
Multi-class classification is a common task that has a well-established conventional loss --- cross-entropy loss.

We use 
accuracy (the percentage of correctly classified images) as our
metric following the common practice.
This task and metric is chosen to demonstrate that for a most common classification setting, UniLoss still performs similarly effectively as the well-established conventional loss.

\subsection{Average Precision for Unbalanced Binary Classification}
\subsubsection{Dataset and Baseline} 
We augment the handwritten digit dataset MNIST to be a binary classification task, predicting zeros or non-zeros. Given an image
containing a single number from 0 to 9, we classify it into the zero (positive) class or
the non-zero (negative) class. The positive-negative ratio of 1:9. 
We create a validation split
by reserving 6k images from the original training set.

We use a 3-layer fully-connected neural network with
500 and 300 neurons in each hidden layer respectively. Our baseline model is trained with
a 2-class cross-entropy loss. 
We train both baseline and UniLoss with a fixed learning rate of 0.01 for 30
epochs. We sample 16 anchors for each anchor type in our anchor interpolation for all of our experiments except in ablation studies.

\subsubsection{Formulation and Refactoring} 
The evaluation process is essentially ranking images using pair-wise comparisons and compute the area under curve based on the ranking. It is determined by whether positive images are ranked higher than negative images. 
Given that the output of a mini-batch of $n$ images is
$\mathbf{s}=(s_1, s_2, \ldots, s_n)$, where $s_i$ represents the predicted score of i-th image
to be positive.  The 
binary variables are 
 $\mathbf{b} = \{b_{i,j} = [c_{i,j} > 0] = [s_i - s_j > 0]\}$, where i belongs to
positives and j belongs to negatives.

\subsubsection{Results}
UniLoss achieves an AP of 0.9988, similarly as the baseline cross-entropy loss (0.9989). 
This demonstrates that UniLoss can effectively optimize for a performance metric (AP) that
is complicated to compute and involves a batch of images.

\subsection{PCKh for Single-Person Pose Estimation}
\subsubsection{Dataset and Baseline}  
We use MPII~\cite{andriluka14cvpr} which has around 22K images for training and 3K images for testing. For simplicity, we perform experiments on the
joints of head only, but our method could be applied to an arbitrary number of human
joints without any modification. 

We adopt the Stacked Hourglass~\cite{newell2016stacked} as our model. The baseline loss is 
the Mean Squared Error (MSE) between the predicted heatmaps and the manually-designed ``ground
truth'' heatmaps. 
We train a
single-stack hourglass network for both UniLoss and MSE using
RMSProp~\cite{hinton2012lecture} with an initial learning rate 2.5e-4 for 30 epochs and then
drop it by 4 for every 10 epochs until 50 epochs.

\subsubsection{Formulation and Refactoring} 
Assume the network generates a mini-batch of heatmaps $\mathbf{s}=(\mathbf{s}^1,\mathbf{s}^2,\ldots,\mathbf{s}^n)\in
\mathbf{R}^{n\times m}$, 
where $n$ is the batch size, $m$ is the number of pixels in each image. 
The pixel with the highest score in each heatmap is predicted as a key point during evaluation.
We note the pixels within the radius $r$ around the ground truth as
positive pixels, and other pixels as negative and each heatmap $\mathbf{s}^k$ can be flatted as 
$(s_{pos,1}^{k}, s_{pos,2}^{k}, \ldots, s_{pos,m_k}^{k},
s_{neg,1}^{k}, \ldots,$
$s_{neg,m-m_k}^{k})$,
where $m_k$ is the number of positive pixels in the k-th heatmap and $s_{pos,j}^{k}$ ($s_{neg,j}^{k}$) is the score of the j-th positive (negative) pixel in the k-th heatmap.

PCKh requires to find out if a positive pixel has the highest score among others. 
Therefore, we need to compare each pair of positive and negative pixels and this leads to the binary variables
$\mathbf{b}=(b_{k,i,j}) \text{ for } 
 1\leq k \leq n$, $1\leq i \leq m_k$, $1\leq j \leq m-m_k$,
 where $b_{k,i,j} = [s_{pos,i}^{k} - s_{neg,j}^{k} > 0]$, i.e. the comparison between the i-th positive pixel and the j-th negative pixel in the k-th heatmap.

\subsubsection{Results} 
It is notable that the manual design of the target heatmaps is a part of the MSE loss function for pose estimation.
It heavily relies on the careful design of the ground truth heatmaps. 
If we intuitively set the pixels at the exact joints to be 1 and the rest of pixels as 0 in the heatmaps, the training diverges. 

Luckily, \cite{tompson2014joint} proposed to design target heatmaps as a 2D Gaussian bump centered on the ground truth joints, whose shape is controlled by its variance $\sigma$ and the bump size. The success of the MSE loss function relies on the choices of $\sigma$ and the bump size. 
UniLoss, on the other hand, requires no such design.

As shown in Table~\ref{tab:pose}, 
our UniLoss achieves a 95.77 PCKh which is comparable as the 95.74 PCKh for MSE with the best $\sigma$. This
validates the effectiveness of UniLoss in optimizing for a pixel-wise prediction problem.

We further observe that the baseline is sensitive to the shape of 2D Gaussian, as in Table~\ref{tab:pose}. Smaller $\sigma$ makes target heatmaps concentrated on ground truth joints and makes the optimization to be unstable. Larger $\sigma$ generates vague training targets and decreases the performance. 
This demonstrates that conventional losses require dedicated manual design while UniLoss can be applied directly.

\begin{table}[!t]
\caption{PCKh of Stacked Hourglass with MSE and UniLoss on the MPII validation.}
\begin{center}
\begin{tabular}{@{}l@{\hskip 0.2in}rrrrrrr@{\hskip 0.2in}r@{}}
\toprule

Loss & MSE $\sigma=0.1$ & $\sigma=0.5 $ & $\sigma=0.7$ & $\sigma=1$ & $\sigma=3$ & $\sigma=5$ & $\sigma=10$ & UniLoss \\ 
\midrule
PCKh & \qquad 91.31 & 95.13 & 93.06 & 95.71 & 95.74 & 94.99 & 92.25 & \textbf{95.77} \\
\bottomrule
\end{tabular}
\end{center}
\label{tab:pose}
\end{table}

\begin{table}[!t]
  \caption{Accuracy of ResNet-20 with CE loss and UniLoss on the CIFAR-10 and CIFAR-100 test set.}
  \begin{center}
  \begin{tabular}{@{}l@{\hskip 1in}r@{\hskip 1in}r@{}}
   \toprule
   Loss & CIFAR-10 & CIFAR-100\\
   \midrule
  CE (cross-entropy) Loss & 91.49 & 65.90\\
   UniLoss  & 91.64 & 65.92\\
   \bottomrule
  \end{tabular}
  \end{center}
  \label{tab:cifar}
\end{table}

\begin{table}[!t]
  \caption{Ablation study for mini-batch sizes on CIFAR-10.}
  \begin{center}
  \begin{tabular}{@{}l@{\hskip 0.3in}rrrrrrrr@{}}
  \toprule
    Batch Size & 8 & 16 & 32 & 64 & 128 & 256 & 512 & 1024\\
    \midrule
    Accuracy & 87.89 & 90.05 & 90.82 & 91.23 & \textbf{91.64} & 90.94 & 89.10 & 87.20 \\
    \bottomrule
  \end{tabular}
  \end{center}
  \label{tlb:batchsize}
\end{table}

\begin{table}[!t]
  \caption{Ablation study for number of anchors on the three tasks.}
  \begin{center}
  \begin{tabular}{@{}l@{\hskip 0.3in}c@{\hskip 0.3in}r@{}}
    \toprule
    Task & \#Anchors in Each Type & Performance \\
    \midrule
     & 5 & 0.9988\\
    Bin. Classification (AP) & 10 & 0.9987\\
     & 16 & 0.9988\\
    \midrule
     & 5 & 91.55\%\\
    Classification (Acc) & 10 & 91.54\%\\
     & 16 & 91.64\%\\
    \midrule
     & 5 & 94.79\%\\
    Pose Estimation (PCKh) & 10 & 94.95\%\\
     & 16 & 95.77\%\\
    \bottomrule
  \end{tabular}
  \end{center}
  \label{tlb:numpoints}
\end{table}

\subsection{Accuracy for Multi-class Classification}

\subsubsection{Dataset and Baseline} 
We use CIFAR-10 and CIFAR-100
~\cite{krizhevsky2009learning}, with $32 \times 32$ images and 10/100 classes. They each have
50k training images and 10k test images. Following prior work~\cite{he2016deep}, we split
the training set into a 45k-5k train-validation split.

We use the ResNet-20 architecture~\cite{he2016deep}. Our
baselines are trained with cross-entropy (CE) loss. All experiments
are trained following the same augmentation and pre-processing techniques as in
prior work~\cite{he2016deep}. 
We use an initial learning rate of 0.1, 
divided by 10 and 100 at the 140th epoch and the 160th epoch, with a total of 200 epochs trained for both baseline and UniLoss on CIFAR-10.
On CIFAR-100, we train baseline with the same training schedule and UniLoss with 5x training schedule because we only train 20\% binary variables at each step. For a fair comparison, we also train baseline with the 5x training schedule but observe no  improvement.

\subsubsection{Formulation and Refactoring} 
As shown in Sec.~\ref{sec:cls_example}, given the output of a mini-batch of n images
$\mathbf{s}=(s_{1,1},s_{1,2}..,s_{n,p})$, we compare the score of the ground truth class and the scores of other $p-1$ classes for each image. That is,
for the i-th image with the ground truth label $y_i$,
$b_{i,j} = [s_{i,y_i} - s_{i,j} > 0]$, where $1\leq j \leq p$, $j\ne y_i$, and $1\leq i\leq n$. 
For tasks with many binary variables such as CIFAR-100, we train a portion of binary variables in each update to accelerate training.

\subsubsection{Results} 
Our implementation of the baseline method obtains a slightly better accuracy (91.49\%) than that was reported
in ~\cite{he2016deep}---91.25\% on CIFAR-10 and obtains 65.9\% on CIFAR-100.
UniLoss performs similarly (91.64\% and 65.92\%) as baselines on both datasets (Table~\ref{tab:cifar}), which shows that even when the conventional loss is well-established for the particular task and metric, UniLoss still matches the conventional loss.

\subsection{Discussion of Hyper-parameters}

\subsubsection{Mini-batch Sizes} 
We also use a mini-batch of
images for updates with UniLoss. 
Intuitively, as long as the batch size is not extremely small or large, 
it should be able to approximate the distribution of the whole dataset. 

We explore different batch sizes on the CIFAR-10 multi-class classification task, as shown in
Table~\ref{tlb:batchsize}. 
The results match with our hypothesis---as long as the batch size is not extreme, the performance is similar. A batch size of 128 gives the best performance.

\subsubsection{Number of Anchors} 
We explore different number of anchors in the three tasks. 
We experiment with 5, 10, 16 as the number of anchors for each type of the good, bad and nearby anchors. That is, we have 15, 30, 48 anchors in total respectively. 
Table \ref{tlb:numpoints} shows that binary classification and classification are less sensitive to the number of anchors, while in pose estimation, fewer anchors lead to slightly worse performance. It is related to the number of binary variables in each task: pose estimation has scores for each pixel, thus has much more comparisons than binary classification and classification. With more binary variables, more anchors tend to be more beneficial.

\section{Conclusion and Limitations}
We have introduced UniLoss, a framework for generating surrogate losses in a unified way, reducing the amount of manual design of task-specific surrogate losses. The proposed framework is based on the observation that there exists a
common refactoring of the evaluation computation for many tasks and performance metrics. 
Using this refactoring we generate
a unified differentiable approximation of the evaluation computation, through
interpolation. 
We demonstrate that using 
UniLoss, we can optimize for various tasks and performance metrics, achieving comparable performance as task-specific losses.

We now discuss some limitations of UniLoss.
One limitation is that the interpolation methods are not yet fully explored. We adopt the most straightforward yet effective way in this paper, such as the sigmoid function and IDW interpolation for simplicity and an easy generalization across different tasks. 
But there are potentially other sophisticated choices for the interpolation methods and for the sampling strategy for anchors. 

The second limitation is that proposed anchor sampling strategy is biased towards the optimal configuration that corresponds to the ground truth when there are multiple configurations that can lead to the optimal performance. 

The third limitation is that ranking-based metrics may result in a quadratic number of binary variables if pairwise comparison is needed for every pair of scores. Fortunately in many cases such as the ones discussed in this paper, the number of binary variables is not quadratic because many comparisons does not contribute to the performance metric.

The fourth limitation is that currently UniLoss still requires some amount of manual effort (although less than designing a loss from scratch) to analyze the given code of the decoder and the evaluator for the refactoring. Combining automatic code analysis with our framework can further reduce manual efforts in loss design. 

\section{Acknowledgements}
This work is partially supported by the National Science Foundation under Grant. 1734266.

\bibliographystyle{splncs04}
\bibliography{egbib}

\section*{Appendix}
\appendix 
\section{Architecture and Training Details}
\subsubsection{Binary Classification} 
The network takes a $28\times28$ image as input and flatten it to be one dimension with 784 elements. It then pass the flattened input through two fully-connected layers, one with 500 neurons and one with 300 neurons. Each fully-connected layer has a ReLU activation layer followed. The prediction is then given with an output layer with a 2-way output.

The baseline loss is a 2-class cross-entropy loss. Our loss is formulated as described in Sec. 4.2. The $h$ function is the sigmoid relaxation of the binary variables. The $g$ function is the interpolation over anchors of the binary variables. For example, the anchor that gives the best performance is that all positive examples have higher scores than negative examples---all $b_{i,j}=1$. Following Sec. 3.4, three types of anchors are generated randomly. The good anchors and nearby anchors are obtained by flipping one binary bit from the best anchor and the anchor at the current training step respectively.We sample 16 anchors for each type.

We train models with the baseline loss and UniLoss with SGD using the same training schedule: with a fixed learning rate of 0.01 for 30 epochs.

\subsubsection{Pose Estimation} 

We use the Stacked Hourglass
Network architecture. 
It takes a 224$\times$224 image as input
and passes it through one hourglass block as described in ~\cite{newell2016stacked}.
It outputs a heatmap for each joint. A heatmap essentially gives scores measuring how likely is the joint to be there for each pixel or region centered at that pixel.

The baseline loss is 
the Mean Squared Error (MSE) between the predicted heatmaps and the manually-designed ``ground
truth'' heatmaps. More specifically, 
the target ``ground truth'' heatmap has  a 2D Gaussian bump centered on the ground truth joint. The shape of the Gaussian bump is controlled by its variance $\sigma$ and the bump size. The commonly chosen $\sigma$ is 1 and the bump size 7.

Our loss is formulated as in Sec. 4.3. The $h$ function is the sigmoid relaxation of the binary variables. The $g$ function is the interpolation over anchors of the binary variables. For example, the anchor that gives the best performance is that all binary values to be 1. Following Sec. 3.4, three types of anchors are generated. 
For good anchors, we flip
a small number of bits from the best. Nearby anchors are flipped from the current configuration by randomly picking a positive/negative pixel in the current output heatmaps and flipping all bits associated with this pixel. We sample 16 anchors for each type. 

We train the model both UniLoss and MSE with
RMSProp~\cite{hinton2012lecture} using an initial learning rate 2.5e-4 for 30 epochs and then
divided by 4 for every 10 epochs until 50 epochs. 

\subsubsection{Classification} 
We use the ResNet-20 architecture~\cite{he2016deep}. 
The network takes a $32\times32$ image as input. The input image first goes through a $3\times3$ convolution layer followed by a batch normalization layer and a ReLU layer. It then goes through three ResNet building blocks with down-sampling in between.  After an average pooling layer, the output layer is a fully-connected layer with a 10-way output.

Our baseline loss is a 10-way cross-entropy (CE) loss. 
Our loss is formulated as in Sec. 4.4. The $h$ function is the sigmoid relaxation of the binary variables. The $g$ function is the interpolation over anchors of the binary variables. For example, the anchor that gives the best performance is that all binary values to be 1, 
meaning that for each image, the ground truth class has
higher score than every other class. 
The good anchors and nearby anchors are obtained by flipping one binary bit from the best anchor and the anchor at the current training step respectively.We sample 16 anchors for each type. 

We use the same data augmentation for both with random cropping with a padding of 4 and random horizontal flipping. We also pre-process the images with per-pixel normalization.
We train both models with SGD using an initial learning rate of 0.1, divided by 10 and 100 at the 140 epoch and the 160 epoch, with a total of 200 epochs on CIFAR-10.
On CIFAR-100, we train baseline with the same training schedule and UniLoss with 5x training schedule but 20\% binary variables at each step. 

\section{Additional Experimental Results}
\subsubsection{Analysis of the Interpolation of $g$}
To evaluate how well the interpolator approximates the true performance metric, we sample binary configurations with various hamming distances from those encountered during training. The hamming distances ranges from 0, $\frac{1}{512}$, …, $\frac{1}{2}$ of total number of binary variables (\#binaries). We compute the L2 distances and rank correlation coefficients between the approximated metric value and true metric value pairs, as in Table~\ref{tlb:hamming}. We see that approximation is quite good when distance is small but poor when distance is large.

\begin{table}[!t]
\caption{The L2 distances and rank correlation coefficients between the approximated metric value and true metric value pairs for binary configurations with various hamming distances from those encountered during training.
}
\begin{center}
\begin{tabular}{@{}l@{\hskip 0.2in}rrrrrr@{}}
\toprule
Hamming Distance (of \#binaries)	& 0	& $\frac{1}{512}$ & $\frac{1}{128}$ &	$\frac{1}{32}$&	$\frac{1}{8}$&	$\frac{1}{2}$\\
\midrule
L2 Distance	& 0.19&	0.19&	0.57&	2.83&	8.49&	12.25\\
Rank Correlation Coeffcient &	0.98&	0.97&	0.92&	0.70&	0.35&	0.15\\
\bottomrule
\end{tabular}
\end{center}
\label{tlb:hamming}
\end{table}

\begin{figure}[!t]
\begin{center}
\includegraphics[width=0.45\linewidth]{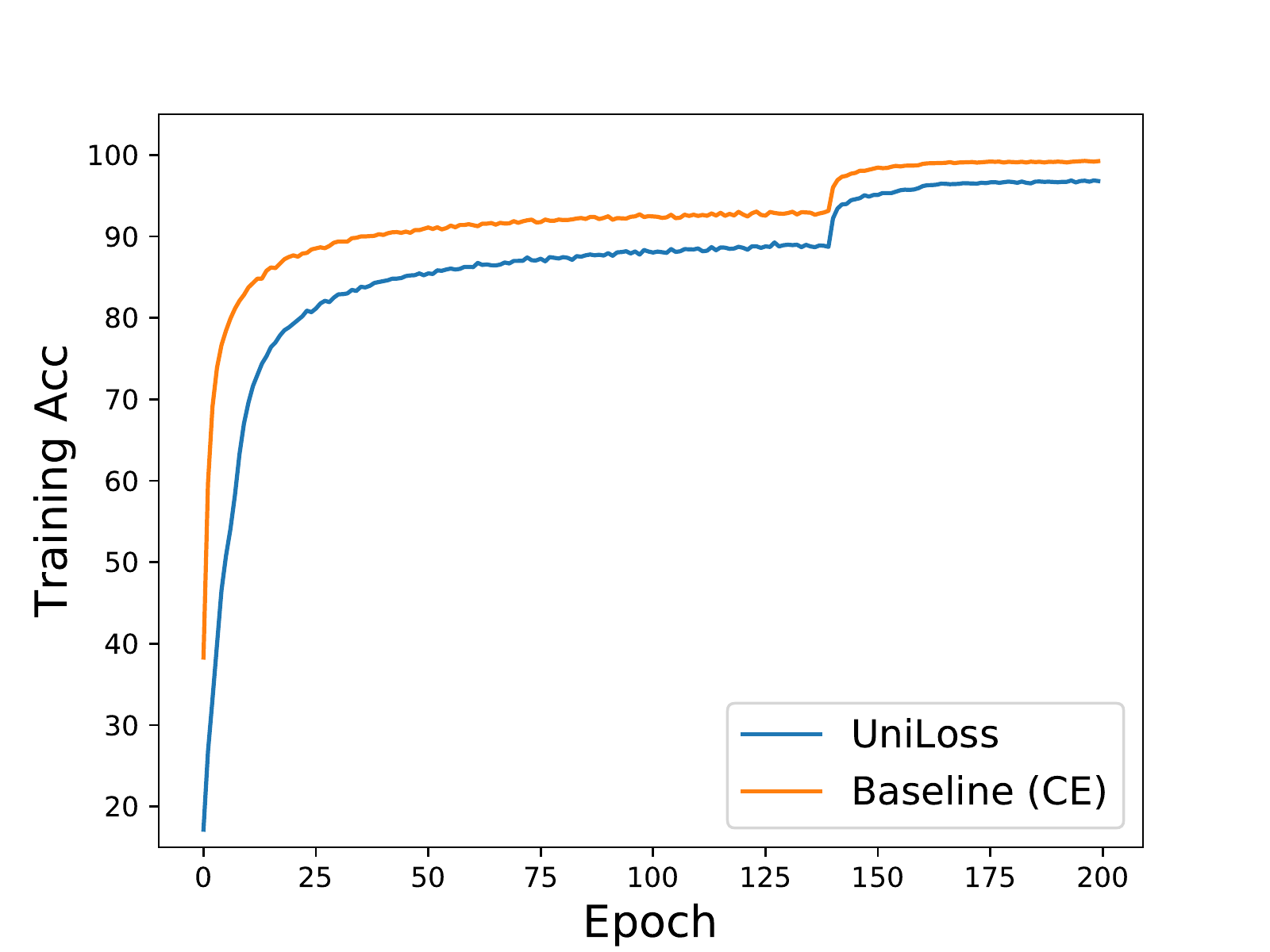}
\includegraphics[width=0.45\linewidth]{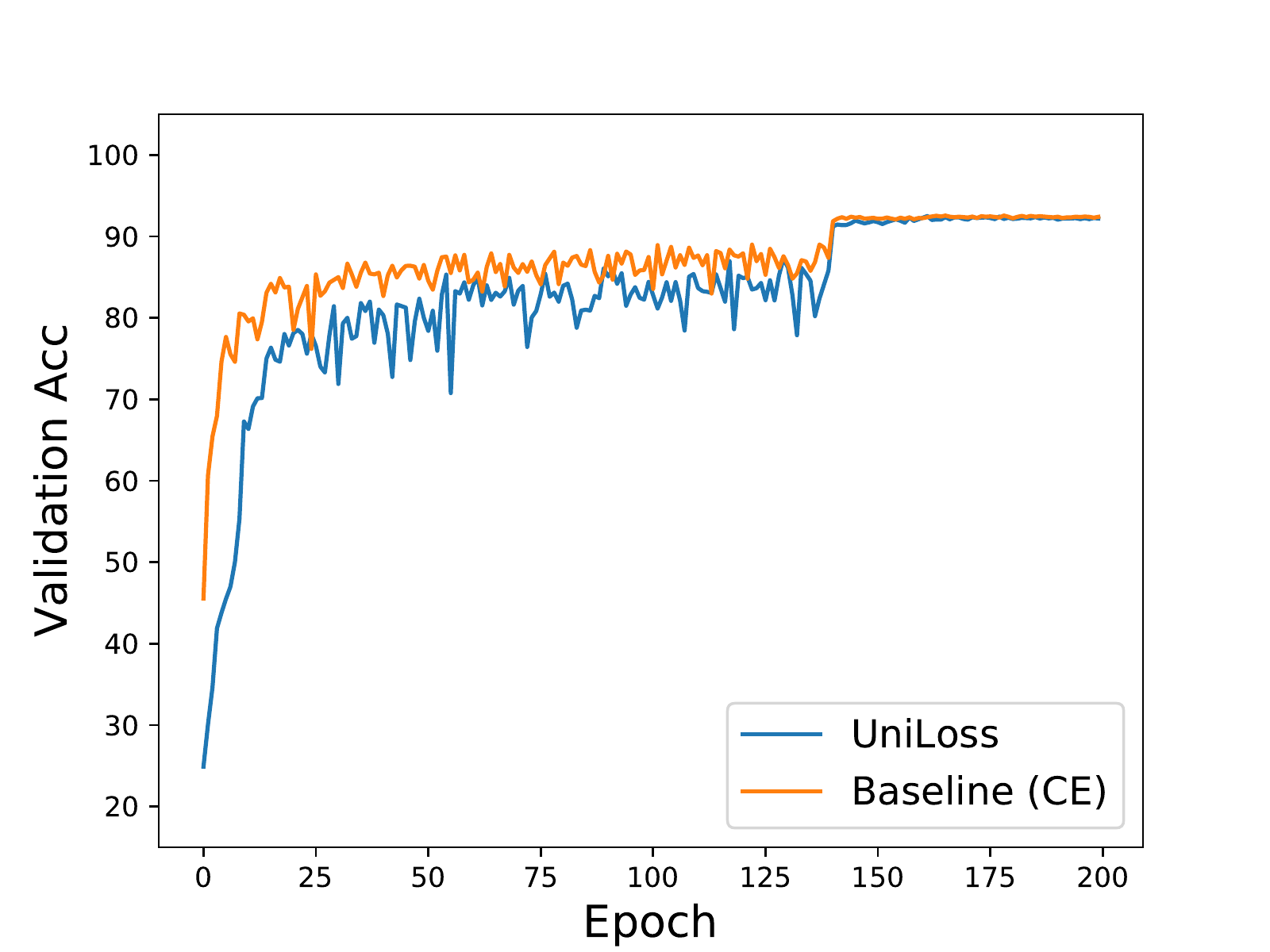}
\end{center}
  \caption{Training and validation accuracy curves for Cross-Entropy (CE) and UniLoss on CIFAR-10.}
\label{fig:cls_curve}
\end{figure}

\begin{figure}[!t]
\begin{center}
\includegraphics[width=0.45\linewidth]{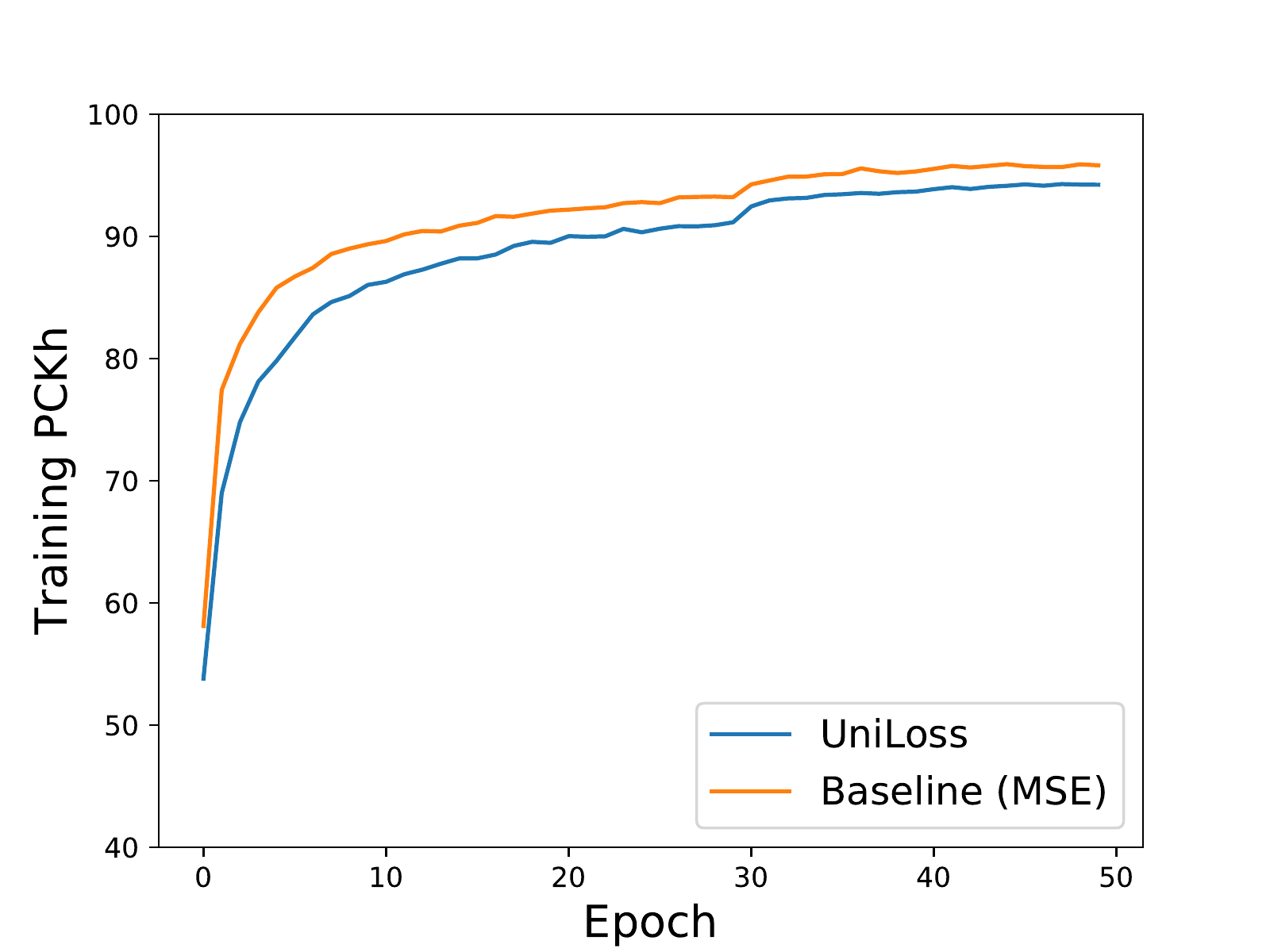}
\includegraphics[width=0.45\linewidth]{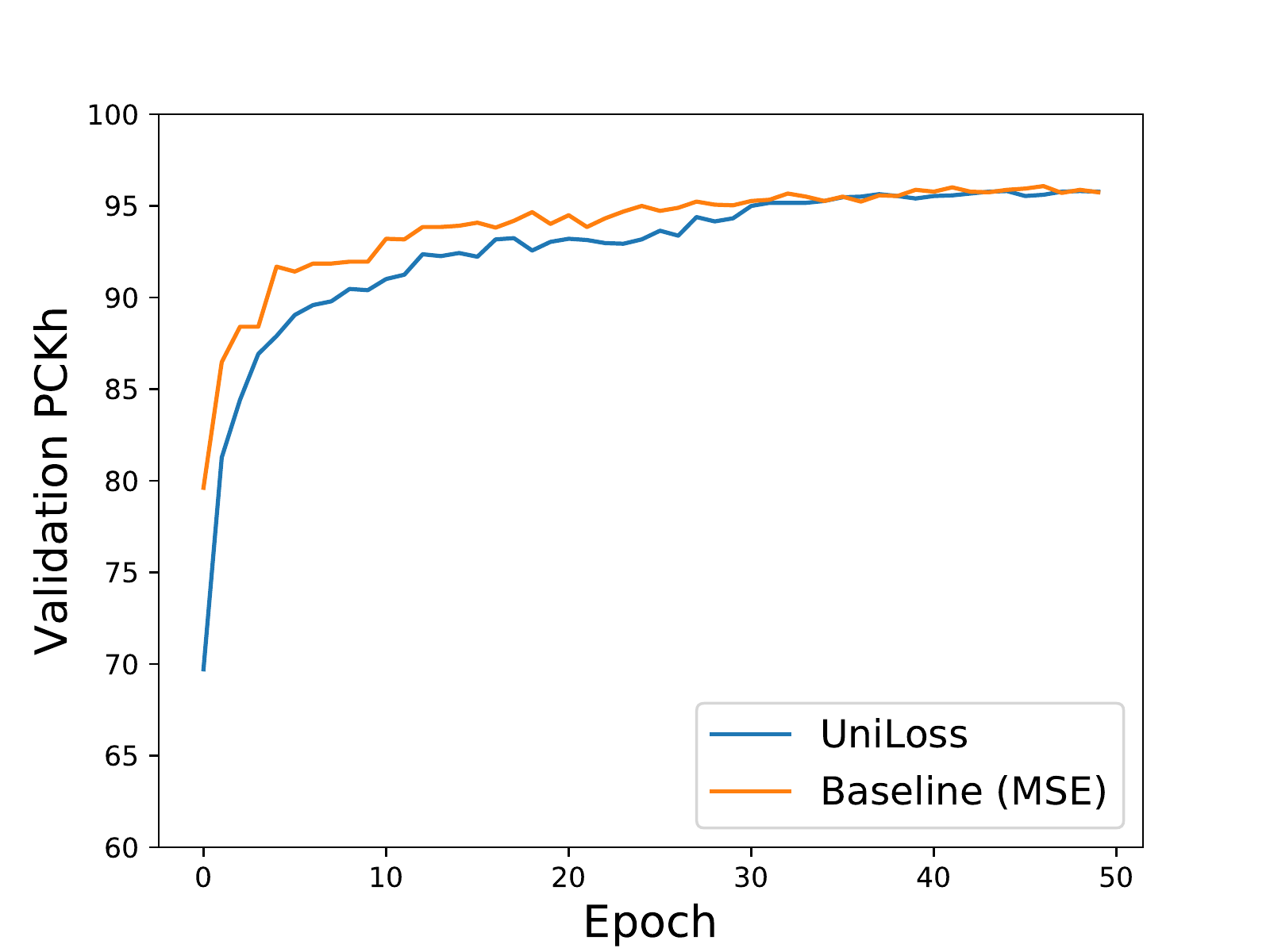}
\end{center}
  \caption{Training and validation PCKh curves for Mean Square Error (MSE and UniLoss on MPII.}
\label{fig:pose_curve}
\end{figure}

\subsubsection{Training and Validation Curves}
We present the training accuracy and validation accuracy curves over epochs in
Fig.~\ref{fig:cls_curve} and Fig. \ref{fig:pose_curve}. 
We see that while UniLoss has similar validation accuracy (PCKh) later in the training with the CE (MSE) baseline, its training accuracy (PCKh) is slightly lower than the baseline over the whole training. 
One hypothesis is that due to the noise 
introduced by the randomness in anchor
 sampling, UniLoss naturally has a regularization effect compared to conventional losses. 
 
\end{document}